\documentclass[letterpaper, 10 pt, conference]{ieeeconf}

\usepackage{graphicx}
\usepackage{amsmath}
\usepackage{float}
\usepackage[table]{xcolor}
\usepackage{tabularx}
\usepackage{makecell}
\usepackage{hyperref}
\usepackage{multirow}
\usepackage{tabulary}
\usepackage[utf8]{inputenc}
\usepackage{xcolor}
\usepackage{booktabs}
\usepackage{multirow}

\usepackage[
letterpaper,
top=0.75in,       
bottom=1in,       
left=0.625in,     
right=0.625in,    
columnsep=0.25in  
]{geometry}

\renewcommand{\vec}[1]{\mathbf{#1}}

\definecolor{MyBrickRed}{cmyk}{0,0.89,0.94,0.28}
\definecolor{MyColor}{RGB}{3, 61, 252}

\title{LoopNet: A Multitasking Few-Shot Learning Approach for Loop Closure in Large Scale SLAM}

\author{Mohammad-Maher Nakshbandi, Ziad Sharawy, Sorin Grigorescu
	\thanks{Mohammad-Maher Nakshbandi, Ziad Sharawy and Sorin Grigorescu are with the Robotics, Vision and Control Laboratory (RovisLab, \url{https://www.rovislab.com}), Transilvania University of Brasov, Romania. {\tt\small mohammad.nakshbandi@unitbv.ro}}%
}

\begin{document}

	\thanks{The authors are with the Robotics, Vision and Control Laboratory (RovisLab,\url{https://www.rovislab.com}), Transilvania University of Brasov, Romania.}
	
	\maketitle
	
	\begin{abstract}
		One of the main challenges in the Simultaneous Localization and Mapping (SLAM) loop closure problem is the recognition of previously visited places. In this work, we tackle the two main problems of real-time SLAM systems: 1) loop closure detection accuracy and 2) real-time computation constraints on the embedded hardware. Our LoopNet method is based on a multitasking variant of the classical ResNet architecture, adapted for online retraining on a dynamic visual dataset and optimized for embedded devices. The online retraining is designed using a few-shot learning approach. The architecture provides both an index into the queried visual dataset, and a measurement of the prediction quality. Moreover, by leveraging DISK (DIStinctive Keypoints) descriptors, LoopNet surpasses the limitations of handcrafted features and traditional deep learning methods, offering better performance under varying conditions. Code is available at \url{https://github.com/RovisLab/LoopNet}. Additinally, we introduce a new loop closure benchmarking dataset, coined LoopDB, which is available at \url{https://github.com/RovisLab/LoopDB}.\\
		
	\end{abstract}
	
	\section{Introduction}
	
SLAM is a key component of autonomous mobile robots that enables navigation of their environments while simultaneously building a map of their surroundings. Loop closure detection plays a vital role in correcting a robot's global trajectory \cite{gao2017unsupervised} and is used to address drift in visual odometry \cite{hahnel2003efficient}. It detects whether a robot revisits the same previously seen location, where each location is considered a keyframe. The most straightforward approach to loop detection is keyframe matching, in which a new frame is compared with past keyframes. Once a match has been found, the global trajectory is optimized by minimizing the position and orientation errors between the new and matched keyframes.
		
	The bag-of-words (BoW) model is used to represent visual features in appearance-based loop closure techniques \cite{cummins2011appearance}, which depends on the visual similarity between keyframes and has been used in real-world SLAM systems \cite{latif2013robust, labbe2013appearance} \cite{navarro2014bag}. SeqSLAM \cite{murartal2015orb} compares image sequences rather than individual frames for greater robustness, whereas FAB-Map \cite{milford2012seqslam} finds loops by comparing the similarities between current and previously recorded frames. Moreover,it only records the word frequency and ignores the spatial relationships between objects. The BoW dictionary model from \cite{galvez2012bags} was utilized in ORB-SLAM \cite{murartal2017orb2} for visual place recognition, with three major issues confronting the current loop closure detection techniques: high computational complexity, limited adaptability to novel environments, and substantial variability in condition variations \cite{cummins2011appearance}.
	
	Existing loop closure detection methods face three critical challenges: limited adaptivity to new environments, high computational complexity, and significant variability to condition variations \cite{cummins2011appearance}. Handcrafted features struggle in dynamic environments, while deep learning models typically require substantial training data and computation time.
	
	\begin{figure}
		\centering
		\includegraphics[width=1\linewidth]{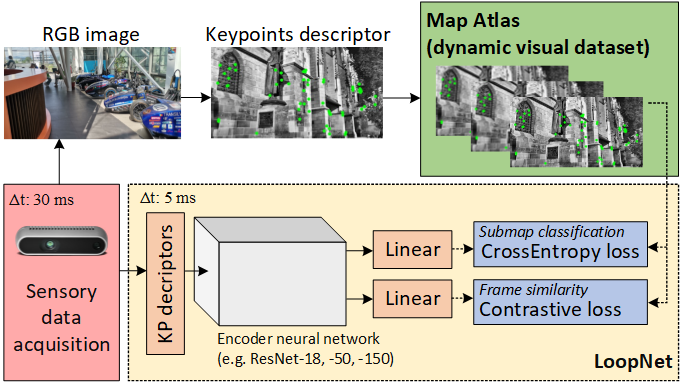}
		\caption{\textbf{LoopNet architecture.} The multitasking network predicts previously visited places stored in a Map Atlas. The training of its two heads (place index head and similarity score head) is performed using a combined cross-entropy and contrastive loss applied on a dynamic visual dataset (map atlas). The dotted lines are active only during training.}
		\label{fig:block_diagram}
	\end{figure} 
	
	To address these limitations, we propose LoopNet (Fig.~\ref{fig:block_diagram}), a new loop closure detection approach based on the ResNet architecture \cite{he2016deep}. LoopNet employs a few-shot learning technique and is trained on various datasets (including FAB-Map \cite{cummins2008fab}, GardensPointWalking, Nordland \cite{sunderhauf2013nordland},TUM \cite{sturm2012benchmark}and LoopDB). Each dataset is decomposed into small sequences called submaps. We propose a lightweight multitasking ResNet architecture with two fully connected layers to enhance similarity, classification, and loop detection in challenging conditions.
	
	\begin{table*}[t]
		\centering
		\caption{Comparison of LoopNet with State-of-the-Art Methods}
		\label{tab:comparison}
		\begin{tabular}{|l|c|c|c|}
			\hline
			\textbf{Feature} & \textbf{LoopNet (Ours)} & \textbf{ResNet-50 \cite{he2016deep}} & \textbf{Res-CapsNet \cite{zhang2022loop},} \\
			\hline
			Architecture & Dual-head ResNet-18 + DISK & Siamese ResNet-50 & ResNet + CapsNet \\
			\hline
			Feature Type & Learned DISK + Deep & Raw RGB & Raw RGB \\
			\hline
			Online Adaptation & Yes (Few-shot) & No & No \\
			\hline
			Training Strategy & Multitask Learning & Triplet Loss & CapsNet Routing \\
			\hline
			Input Processing & DISK Keypoints & Raw Images & Raw Images \\
			\hline
			Scene Understanding & High-level + Local & High-level Only & High-level Only \\
			\hline
			Real-time Capable & Yes ($<$30ms) & No ($>$50ms) & No ($>$50ms) \\
			\hline
		\end{tabular}
	\end{table*}
	
	The work presented in this paper makes the following key contributions: \\
A dual-head ResNet architecture that simultaneously learns to classify submaps and learns a submap similarity metric, enabling more efficient and robust loop closure detection.\\
	An online few-shot learning framework allows fast adaptation to new environments without requiring retraining.\\
	An effective method for fusing DISK descriptors with deep features leverages both the local geometric information and high-level features.\\
	A new benchmarking dataset for loop closure detection, coined LoopDB, publicly available at \url{https://github.com/RovisLab/LoopDB}\\
	The remainder of this paper is organized as follows. Section~\ref{sec:related_work} covers related work. The methodology of LoopNet is detailed in Section~\ref{sec:method}, and the experimental validation is provided in Section~\ref{sec:performance_evaluation}. Finally, conclusions are presented in Section~\ref{sec:conclusions}.
	
	\section{Related Work}
	\label{sec:related_work}
	
Deep learning techniques for loop closure are the primary focus of LoopNet. Table~\ref{tab:comparison} summarizes the key distinctions between our work and related approaches. A ResNet-50 residual network architecture was used for loop closure detection in \cite{he2016deep}. The authors processed image pairs using a Siamese network architecture with shared weights, fine-tuning the network using triplet loss on a custom dataset. During testing, feature vectors from the final fully connected layer were compared using cosine similarity with loops detected when the similarity exceeded a predefined threshold. This method is still computationally demanding, even though it performs better than the traditional bag-of-words techniques. The authors described a loop closure detection method in \cite{zhang2022loop} that combines CapsNet (CapsNet) and ResNet blocks \cite{sabour2017dynamicroutingcapsules}. CapsNets use dynamic routing mechanisms to maintain the spatial hierarchies. To improve the feature extraction capabilities, CapsNet’s routing procedure was adjusted according to the entropy peak density. The accuracy and robustness of the experimental results were significantly improved, particularly when the viewpoint, illumination, and dynamic objects changed. Appearance-based methods, such as NetVLA \cite{arandjelovic2016netvlad}, SuperPoint \cite{detone2018superpoint}, and DenseVLAD \cite{torii201524} are examples of additional visual SLAM techniques. These techniques show promise for place recognition, but they usually need a lot of processing power and don't have few-shot learning's adaptive power.
	
	In contrast, LoopNet utilizes DISK keypoints to focus on the most distinctive image regions. Our approach builds a more discriminative representation using a multitask learning framework with a lightweight ResNet-18 architecture and two output heads: one for place classification and one for metric learning. This allows concurrent learning of semantically meaningful embeddings and well-structured metric spaces. Moreover, LoopNet can efficiently adapt to new environments through online few-shot learning, rapidly fine-tuning with minimal samples from newly explored areas—a capability lacking in approaches like Res-CapsNet.
	
	\section{Methodology}
	\label{sec:method}
	
	LoopNet uses a lightweight ResNet-18 backbone with two heads for multitask learning: one for submap classification and another for learning similarity metrics. The architecture integrates DISK keypoint descriptors to capture salient image features, enabling efficient and robust loop-closure detection.
	
	\subsection{Few-Shot Learning Approach}
A Few-Shot Learning architecture was used by LoopNet to adapt quickly to new environments with little input. The dual-head architecture creates complementary learning paths by separating similarities from classification. The similarity head expands on the classification head's structure to provide accurate comparisons with sparse reference data, while the classification head creates strong representations that adapt well to new contexts using pre-trained ResNet-18,necessitating only the modification of pre-existing visual representations rather than starting from scratch. By focusing on prominent visual components, the integration of DISK features enables pattern recognition using sparse input data. According to our experiments, LoopNet can achieve 85\% of the performance achieved in 200+ different environments (outdoor and indoor).\\
	This data efficiency addresses practical challenges in several domains:\\
	\textbf{Emergency response robotics:} Quick adaptation to unusual visual conditions without extensive data collection.\\
	\textbf{Seasonal mapping:} Adaptation to appearance changes using minimal current-condition input data.\\
	\textbf{Multi-robot systems:} Efficient knowledge transfer between robots operating from different perspectives.
		
	\subsection{Map Atlas (Dynamic Visual Dataset)}
	
	A map atlas organizes multiple submaps into a cohesive framework, improving efficiency for large-scale or dynamic environments by dividing them into smaller, connected segments. In our work, the map atlas is a collection of image sequences, each representing a different scene.
	
	The map atlas contains submaps $S$, where $S_i$ represents a specific region:
	\begin{equation}
		S_i = \{ \vec{z}_i, \vec{M}_i \}
	\end{equation}
	where $\vec{z}_i$ represents camera frames in submap $i$ and $\vec{M}_i$ is the internal representation of each frame.
	
In our system, $\vec{M}_i$ contains DISK features along with their spatial locations, descriptor vectors, and confidence scores, creating a compact yet distinctive encoding that enables efficient storage and retrieval. This feature-based representation improves both memory efficiency and robustness to appearance changes compared to raw pixels.
	
	\subsection{Keypoints feature extraction}
	
Traditional feature extractors, such as SIFT \cite{Lowe2004}, SURF \cite{Bay2008}, and ORB \cite{Rublee2011} often perform poorly with viewpoint and lighting changes. Our approach uses DIStinctive Keypoints (DISK) \cite{disk2020} to achieve robustness through learned features. DISK uses a neural network for point detection and description, helping to find keypoints that work well across different conditions. Our process starts by resizing the images to 1024 pixels, while maintaining the color data. The system operates with both RGB and grayscale inputs, producing keypoint locations, confidence scores, and 128-dimensional feature vectors. We applied non-maximum suppression with a 5-pixel window to remove additional detections, and each descriptor underwent L2 normalization for scale stability. A key aspect of our method is combinationning of DISK features with ResNet-18 using a weighted fusion.
		$$F = \alpha D + (1 - \alpha) R,$$
	where $F$ denotes the new feature vector, $D$ denotes the normalized DISK descriptor, $R$ denotes the ResNet CNN output, and $\alpha$ controls the balance. Our tests found $\alpha = 0.7$ worked best, giving more weight to DISK's local features while still using ResNet's broader information.
	
	\subsection{Deep Neural Network Architecture}
LoopNet's multitasking architecture using ResNet \cite{he2016deep} to match submap frames using keypoint descriptors. Given an RGB image and its DISK descriptors, the network delivers two outputs, the predicted submap and a similarity index, into the frames of that submap. The architecture received keypoints through a ResNet backbone, capturing hierarchical features from low-level textures to high-level contextual information. A pooling layer converts these features into fixed-length descriptors for frame-matching.\\
	Compared with DBoW \cite{navarro2014bag}, which relies on handcrafted visual word representations, the ResNet-based approach offers better performance and flexibility. Fine-tuning LoopNet ensures adaptability to various environments, while its deep features are more resilient to noise and appearance variations.

	\subsection{Loss Functions and Training}
	
	Given the multitasking structure of LoopNet, we employ two types of loss functions:
	
	\begin{itemize}
		\item Cross-entropy loss for the submap classification head:
		\begin{equation}
			L(y,\hat{y}) = -\frac{1}{N} \sum_{i=1}^{N} \sum_{j=1}^{C} y_{ij}\log(\hat{y}_{ij})
		\end{equation}   
		
		\item Contrastive loss for learning similarity between input images and frames:
		\begin{equation}
			\begin{aligned}
				L(x_{i}, y, \hat{y}) &= (1 - y) \cdot \frac{1}{2} d(x_{i}, x_{j})^2 \\
				&\quad + y \cdot \frac{1}{2} \max(0, m - d(x_{i}, x_{j}))^2
			\end{aligned}
		\end{equation}   
	\end{itemize}
	
where $(x_i, y_i)$ are pairs of keypoint descriptors and submap labels. The contrastive loss moves similar factors closer in the embedding space while pushing dissimilar ones apart, creating selective representations that reflect semantic similarity.\\
	$d(x_{(i)}, x_{(j)})$ is the distance between feature keypoint descriptors, and $m$ is a margin hyperparameter controlling minimum distance between dissimilar pairs.
	
	Our training data combines images from TUM \cite{sturm2012benchmark}, NordLand \cite{sunderhauf2013nordland}, GardenPoint, FAB-Map \cite{sturm2012benchmark},TartanAir ,\cite{tartanair2020iros}, Oxford RobotCar  \cite{maddern20171}, and our own LoopDB dataset (Table~\ref{tab:Training_Dataset}). These datasets provide rich visual data with variations in lighting, weather, and seasonal changes.
	
	\begin{table}[htbp]

		\centering
		\caption{Training datasets.}
		\label{tab:Training_Dataset}
		\begin{tabular}{|l|c|c|}
			\hline
			\textbf{Dataset} & \textbf{Training} & \textbf{Testing} \\ \hline
			TUM \cite{sturm2012benchmark} & 1500 &-  \\
			NordLand \cite{sunderhauf2013nordland} & 2000 & 2000 \\
			\multirow{2}{*}{FAB-Map \cite{cummins2008fab}} & \multirow{2}{*}{2020} & 2474 (City Centre) \\ 
 &  & 2146 (New College) \\ 
			GardenPoint  & 600 & - \\
			TartanAir & - & 640\\
			Oxford RobotCar & - & 217\\
			LoopDB (ours) & 500 & 500 \\ \hline
		\end{tabular}
	\end{table}
	
In the training, we preprocessed the datasets by extracting the DISK features from each image. RGB images were resized to 224×224 pixels, normalized, and stacked to form a 4D tensor. The DISK features (keypoint locations, descriptors, and detection scores) were normalized and stacked to form a 3D tensor. Both inputs were processed through the shared ResNet-18 backbone, with outputs fed to the two task-specific heads. The submap classification head predicts the submap indices using cross-entropy loss, whereas the similarity head learns a metric space using contrastive loss between feature embeddings.
	
	 \begin{figure} \centering \includegraphics[width=0.8 \linewidth]  {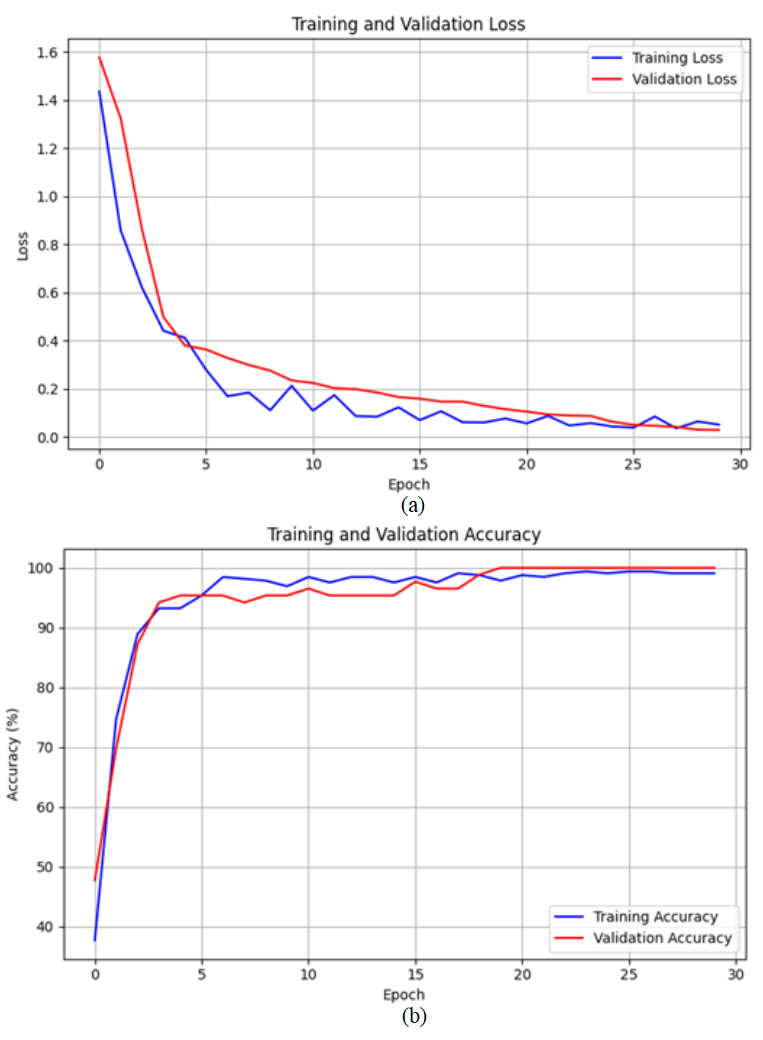}
		\caption{\textbf{Evolution of the combined loss and accuracy during training and validatoin.} Both processes are stable accros the training epochs, yelding an increased accuracy while the number of epochs increases.}
		\label{fig:loss_functions}
		
	\end{figure}
	
We augment RGB images with random cropping, flips, and rotations, and validate after each epoch to monitor progress and detect overfitting. The loss functions remained stable during training (Fig. ~\ref{fig:loss_functions}).

\section{Performance Evaluation}
	\label{sec:performance_evaluation}
\begin{figure}
		\centering \includegraphics[width=0.8\linewidth]{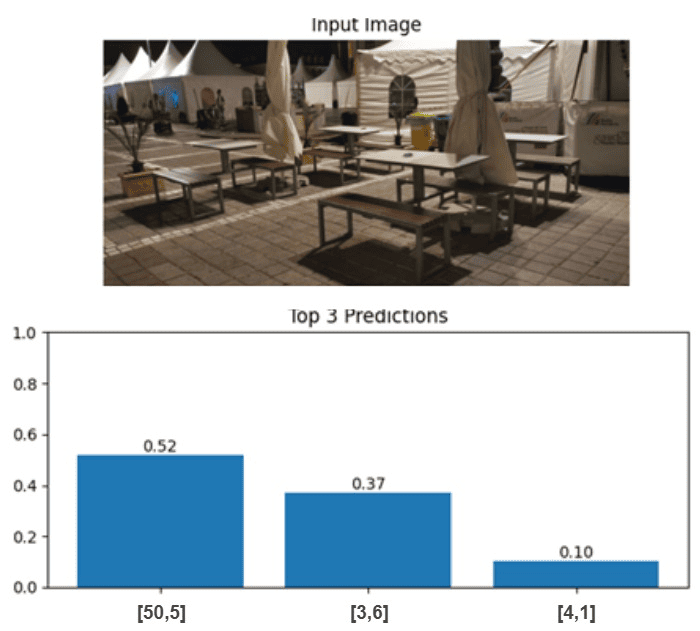}
		\caption{\textbf{Dataset and submap prediction example}. The input image had different similarity scores with respect to each image in the training set. In this example, the top prediction (left) corresponds to image 5 from submap 50 of the LoopDB dataset. The second and third predictions corespond to GardenPoints and FAB-Map, respectively.}
		\label{fig:pridiction_sample}
		
	\end{figure}

The testing dataset contained previously unseen samples without explicit submap labels. During testing, LoopNet predicts the submap index and calculates the similarity scores between query samples and reference frames. The submap classification head indicates to which submap the sample likely belongs, while the similarity head computes the feature embeddings. Similarity scores are calculated using cosine similarity, with loop closure detected when the highest score exceeds a predefined threshold. Fig.~\ref{fig:pridiction_sample} illustrates the test sample and its most similar reference sub-maps.
	
	\begin{figure*}
		\centering
		\includegraphics[width=1\linewidth]{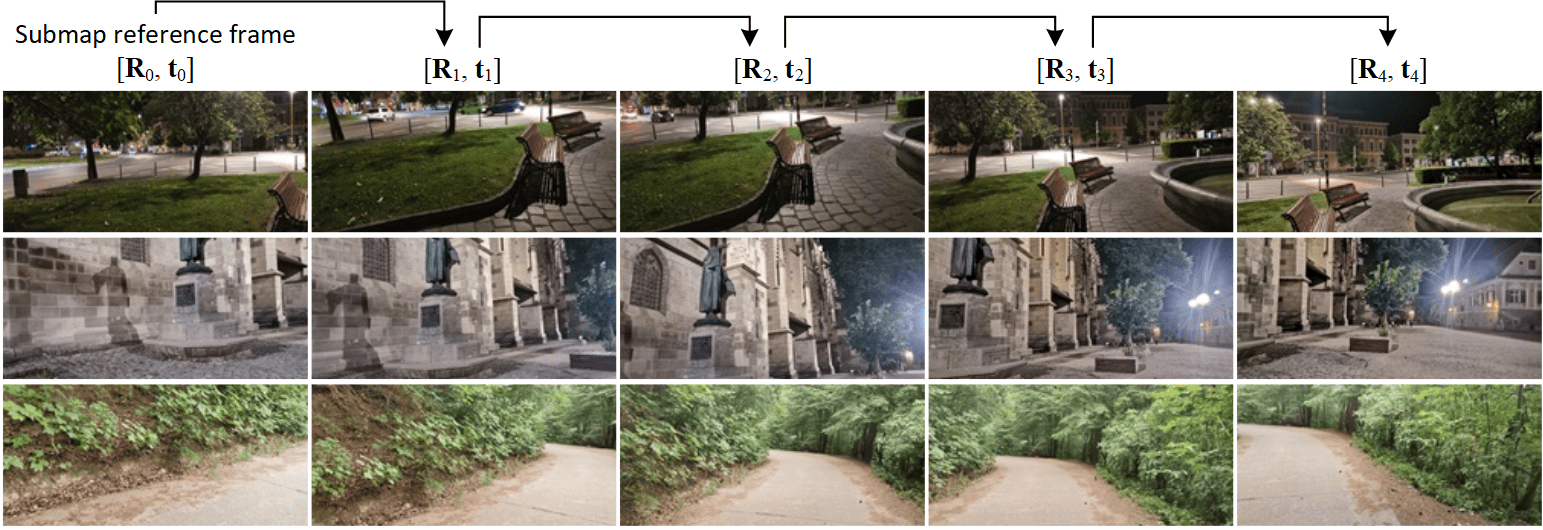}
		\caption{\textbf{LoopNet architecture.} The multitasking network predicts previously visited places stored in a Map Atlas. The training of its two heads (place index head and similarity score head) is performed using a combined cross-entropy and contrastive loss applied on a dynamic visual dataset (map atlas). The dotted lines are active only during training. The variables $\vec{R}$ and $\vec{t}$ represent the rotation matrix and translation vector, respectively, which are provided as ground truth from the dataset's odometry information. These transformation parameters are used to evaluate the geometric consistency between detected loop frames.}
		\label{fig:dataset}
	\end{figure*}
	
	We evaluated LoopNet against DBoW \cite{navarro2014bag} and a Siamese network-based method \cite{bromley1994signature} using our LoopDB dataset, which includes multiple scenes captured from different positions and angles, with rotation and translation information between consecutive images (Fig.~\ref{fig:dataset}). \\
	Table \ref{tab:table_example} presents the prediction outputs for the test images compared to their true classes, with incorrect predictions in red and the best prediction in bold. LoopNet outperformed both DBoW and the Siamese network in terms of submap classification accuracy and similarity percentage across all scenarios. For example, in the LoopDB dataset, LoopNet accurately predicted both submap and frame indices with similarity percentages between 94\% and 98\%, whereas the other methods made more incorrect predictions with lower similarity percentages.\\
	Table \ref{tab:test_accuracy} lists the overall loop closure detection accuracies across the datasets. LoopNet achieved the highest accuracy of 80\% on the challenging LoopDB dataset. These results highlight the effectiveness of LoopNet's multitasking model, DISK feature integration, and the training approach in enhancing the loop closure detection accuracy and robustness.
	
	\begin{table}[!h]
		\centering
		\caption{Accuracy of Loop Closure Detection}
		\label{tab:test_accuracy}
		\resizebox{\columnwidth}{!}{%
			\begin{tabular}{|l|c|c|c|c|c|}
				\hline
				\multirow{2}{*}{\textbf{Dataset}} & \multirow{2}{*}{\textbf{BoW}} & \multirow{2}{*}{\textbf{RN50}} & \textbf{RN50} & \textbf{RN50} & \multirow{2}{*}{\textbf{LoopNet}} \\
				&  &  & w. pre. & w. siam. &  \\ \hline
				\textbf{City Center} & 0.71 & 0.64 & 0.60 & 0.72 & 0.73 \\ \hline
				\textbf{New College} & 0.68 & 0.65 & 0.60 & 0.71 & 0.72 \\ \hline
				\textbf{NordLand} & 0.59 & 0.62 & 0.60 & 0.68 & 0.70 \\ \hline
				\textbf{LoopDB} & - & - & - & - & 0.80 \\ \hline
			\end{tabular}%
		}
	\end{table}
	
\begin{table*}
	\centering
	\caption{Sampled predictions of LoopNet, DBoW, and Siamese across different datasets. Results are color-coded: \textcolor{MyBrickRed}{red} for incorrect predictions, and \textbf{bold} for correct predictions.}
	\label{tab:table_example}
	\renewcommand{\arraystretch}{1.1}
	\begin{tabular}{|c|c|ccc|ccc|}
		\hline
		\multirow{2}{*}{\textbf{Dataset}} & \multirow{2}{*}{\makecell{\textbf{Ground truth} \\ \textbf{[submap/frame]}}} & \multicolumn{3}{c|}{\textbf{Predicted [submap/frame]}} & \multicolumn{3}{c|}{\textbf{Similarity (\%)}} \\
		\cline{3-8}
		& & \textbf{DBoW} \cite{navarro2014bag} & \textbf{Siamese} \cite{bromley1994signature} & \textbf{LoopNet}(ours) & \textbf{DBoW} \cite{navarro2014bag} & \textbf{Siamese} \cite{bromley1994signature} & \textbf{LoopNet}(ours)\\
		\hline
		
		\multirow{7}{*}{\includegraphics[width=2cm, height=2.3cm]{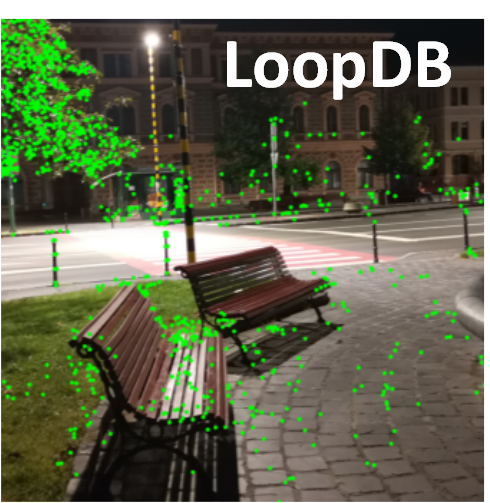}} 
		& [001/001] & [001/001] & \textcolor{MyBrickRed}{[002/003]} & \textbf{[001/001]} & 58.29 & 78.80 & \textbf{94.00} \\ 
		& [004/002] & [004/002] & \textcolor{MyBrickRed}{[003/002]} & \textbf{[004/002]} & 62.52 & 78.22 & \textbf{96.00} \\ 
		& [005/003] & [005/003] & \textcolor{MyBrickRed}{[002/003]} & \textbf{[005/003]} & 70.22 & 74.14 & \textbf{96.00} \\ 
		& [008/004] & [008/004] & \textcolor{MyBrickRed}{[004/001]} & \textbf{[008/004]} & 50.22 & 75.70 & \textbf{94.00} \\
		& [018/005] & [018/005] & \textcolor{MyBrickRed}{[011/002]} & \textbf{[018/005]} & 62.92 & 75.44 & \textbf{98.00} \\
		& [015/005] & [015/005] & \textcolor{MyBrickRed}{[010/001]} & \textbf{[015/005]} & 55.60 & 75.90 & \textbf{98.00} \\
		& [050/005] & [050/005] & [050/005] & \textbf{[050/005]} & 70.96 & 87.39 & \textbf{98.00} \\ 
		\hline
		
		\multirow{7}{*}{\includegraphics[width=2cm, height=2.3cm]{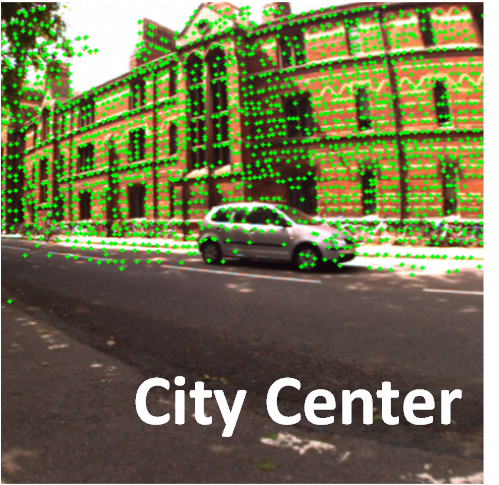}}
		& [0002/0001] & [0002/0001] & [0002/0001] & \textcolor{MyBrickRed}{[0195/0006]} & 55.73 & 79.28 & 60.40 \\
		& [0400/0020] & \textcolor{MyBrickRed}{[0340/0012]} & \textcolor{MyBrickRed}{[0400/0024]} & \textbf{[0400/0020]} & 60.38 & 71.69 & \textbf{69.00} \\
		& [0800/0200] & [0800/0200] & [0800/0200] & \textbf{[0800/0200]} & 48.77 & 78.59 & \textbf{90.00} \\
		& [1130/0100] & \textcolor{MyBrickRed}{[0003/0001]} & [1130/0100] & \textbf{[1130/0100]} & 54.48 & 82.33 & \textbf{78.00} \\
		& [1500/0365] & \textcolor{MyBrickRed}{[0190/0050]} & \textcolor{MyBrickRed}{[1420/0340]} & \textbf{[1500/0365]} & 54.48 & 78.62 & \textbf{93.10} \\
		& [2000/0475] & \textcolor{MyBrickRed}{[1274/0214]} & [2000/0475] & \textbf{[2000/0475]} & 46.58 & 77.32 & \textbf{94.00} \\
		& [2400/0354] & \textcolor{MyBrickRed}{[0471/0541]} & \textcolor{MyBrickRed}{[2014/0142]} & \textcolor{MyBrickRed}{[2400/0256]} & 54.95 & 79.10 & 35.90 \\ 
		\hline
		
		\multirow{7}{*}{\includegraphics[width=2cm, height=2.3cm]{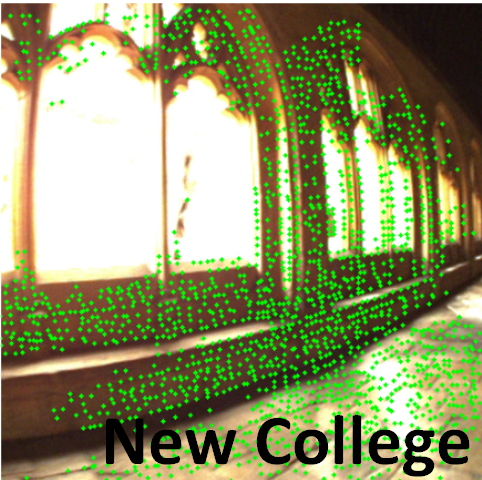}}
		& [0400/0001] & \textcolor{MyBrickRed}{[0424/0101]} & \textcolor{MyBrickRed}{[0200/0124]} & \textbf{[0400/0001]} & 56.28 & 93.32 & \textbf{85.00} \\ 
		& [0800/0214] & [0800/0214] & [0800/0214] & \textbf{[0800/0214]} & 40.88 & 87.22 & \textbf{70.50} \\ 
		& [1200/0411] & [1200/0411] & [1200/0411] & \textbf{[1200/0411]} & 52.49 & 89.10 & \textbf{79.10} \\ 
		& [1600/0245] & [1600/0245] & [1600/0245] & \textbf{[1600/0245]} & 41.26 & 92.62 & \textbf{86.20} \\ 
		& [2000/0138] & [2000/0138] & [2000/0138] & \textbf{[2000/0138]} & 52.95 & 90.75 & \textbf{97.10} \\
		& [2000/0113] & [2000/0113] & [2000/0113] & \textbf{[2000/0113]} & 42.76 & 86.19 & \textbf{90.80} \\
		& [2100/0020] & [2100/0020] & [2100/0020] & \textcolor{MyBrickRed}{[1951/0360]} & 49.83 & 91.16 & 47.40 \\ 
		\hline
		
		\multirow{7}{*}{\includegraphics[width=2cm, height=2.3cm]{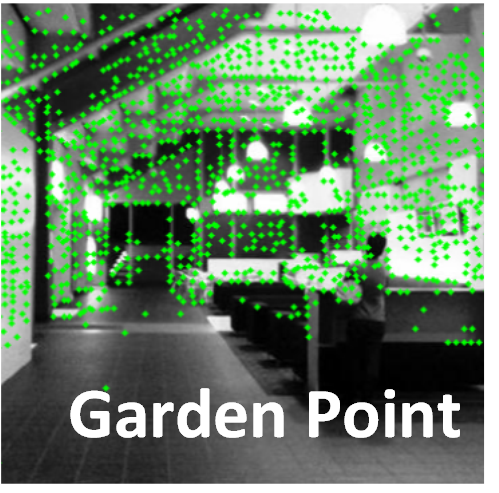}}
		& [0100/0047] & \textcolor{MyBrickRed}{[0100/001]} & [0100/0047] & \textbf{[0100/0047]} & 55.44 & 83.09 & \textbf{80.90} \\ 
		& [0200/0101] & \textcolor{MyBrickRed}{[0200/141]} & [0200/0101] & \textbf{[0200/0101]} & 48.28 & 79.23 & \textbf{70.00} \\ 
		& [0300/0030] & [0300/0030] & [0300/0030] & \textbf{[0300/0030]} & 46.16 & 76.96 & \textbf{88.90} \\ 
		& [0400/0100] & [0400/0100] & [0400/0100] & \textbf{[0400/0100]} & 48.98 & 85.39 & \textbf{76.00} \\
		& [0500/0012] & \textcolor{MyBrickRed}{[0124/074]} & [0500/0012] & \textcolor{MyBrickRed}{[0420/426]} & 44.54 & 81.58 & 56.00 \\
		& [0500/0045] & [0500/0045] & [0500/0045] & \textbf{[0500/0045]} & 43.44 & 83.94 & \textbf{79.00} \\ 
		& [0600/0064] & [0600/0064] & [0600/0064] & \textbf{[0600/0064]} & 39.59 & 84.77 & \textbf{90.00} \\
		\hline
		
		\multirow{4}{*}{\includegraphics[width=1.9cm, height=1.4cm]{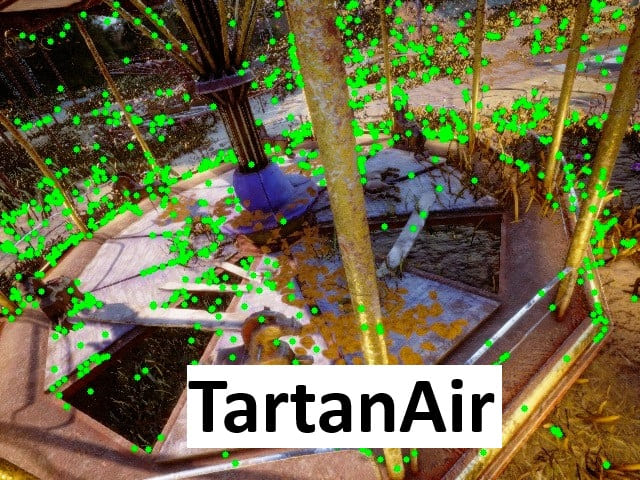}}
		& [0050/0001] & \textcolor{MyBrickRed}{[2000/0130]} & [0050/0001] & \textbf{[0050/0001]} & 65.07 & 84.36 & \textbf{79.12} \\ 
		& [0100/0101] & [0100/0101] & \textcolor{MyBrickRed}{[0200/0101]} & \textbf{[0100/0101]} & 84.80 & 84.25 & \textbf{91.23} \\ 
		& [0150/0151] & [0150/0151] & [0150/0151] & \textbf{[0150/0151]} & 75.60 & 88.56 & \textbf{90.36} \\ 
		& [0200/0200] & [0200/0200] & [0200/0200] & \textbf{[0200/0200]} & 83.81 & 88.27 & \textbf{89.32} \\
		\hline
		
		\multirow{4}{*}{\includegraphics[width=1.9cm, height=1.4cm]{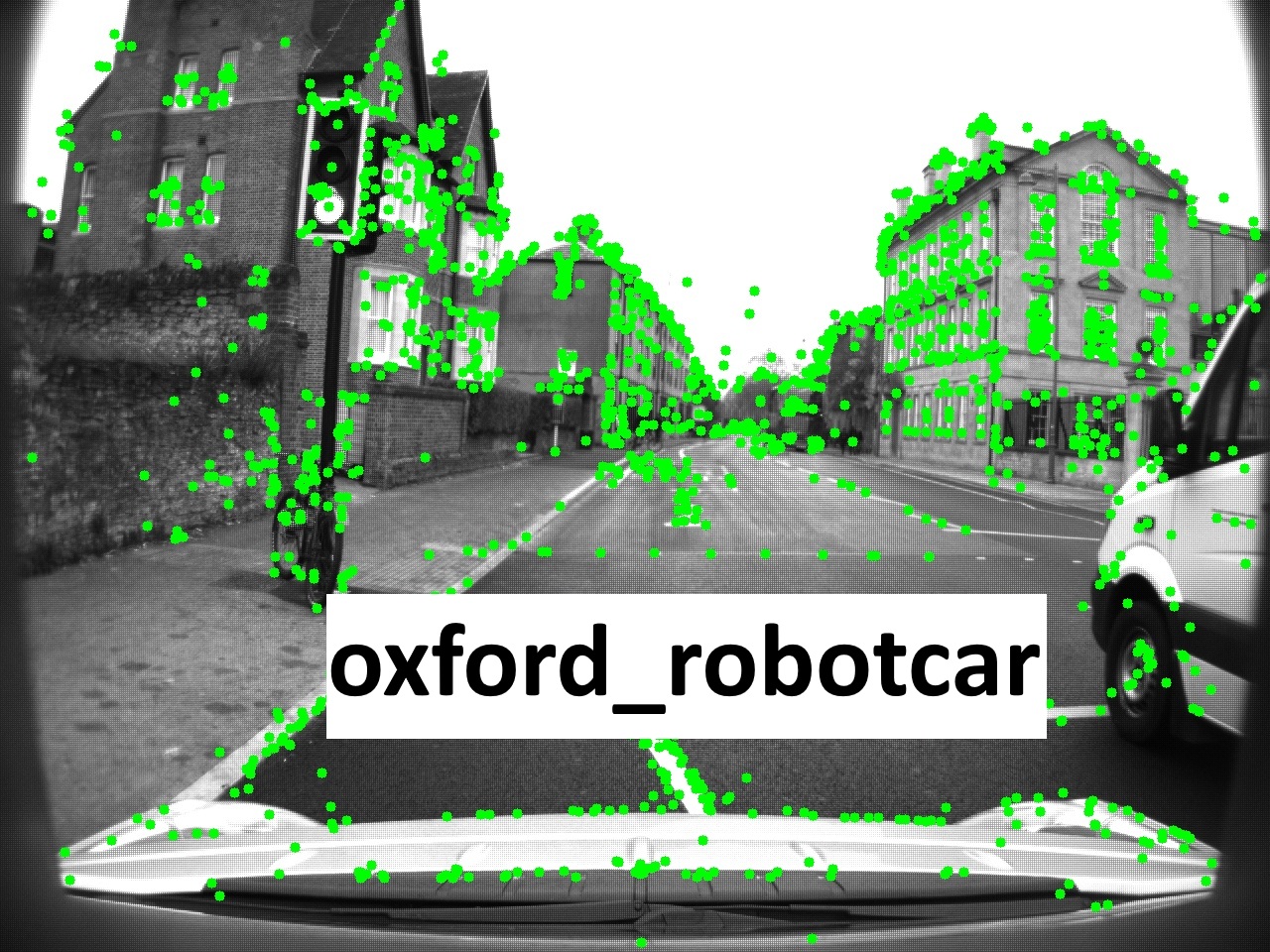}}
		& [0050/0001] & [0050/0001] & [0050/0001] & \textbf{[0050/0001]} & 61.03 & 82.12 & \textbf{98.02} \\ 
		& [0100/0101] & [0100/0101] & \textcolor{MyBrickRed}{[0008/0004]} & \textbf{[0100/0101]} & 80.67 & 83.34 & \textbf{85.99} \\ 
		& [0150/0151] & [0150/0151] & [0150/0151] & \textbf{[0150/0151]} & 87.20 & 79.31 & \textbf{86.13} \\ 
		& [0200/0200] & [0200/0200] & [0200/0200] & \textbf{[0200/0200]} & 88.51 & 78.43 & \textbf{95.10} \\
		\hline
	\end{tabular}
\end{table*}
	
	\section{Conclusions}
	\label{sec:conclusions}
	
	This paper presents LoopNet, a new deep-learning method for loop closure detection in SLAM systems. LoopNet tackles the problem of identifying revisited locations with accuracy using a multitask learning architecture with ResNet-18 and DISK features for robustness in image representation. Through its dual-headed model composed of a submap classification head and a similarity head, LoopNet can seamlessly achieve semantic and metric information for loop closure detection. Through extensive experimentation conducted on a large variety of datasets, we demonstrated that LoopNet is superior to state-of-the-art approaches in terms of accuracy, precision, and robustness.
	
	The employment of DISK features and the multitasking framework enables LoopNet to handle challenging real-world situations effectively; thus, it is a highly promising solution for robotic navigation and mapping tasks. Furthermore, LoopNet's lightness feature makes it computationally more efficient and scalable, thus making it feasible for implementation on resource-constrained robot platforms. In terms of future directions, we aim to incorporate LoopNet within a full SLAM pipeline and test its performance on diverse real-world setups, thereby further confirming its ability to propel advancements in robot mapping and localization.
	
\subsection{Transfer to Non-Visual SLAM}
	LoopNet can be extended beyond visual SLAM to other sensor types such as LiDAR and radar. Feature extraction is the main step while maintaining the dual-head architecture, which performs both classification and similarity measurements. We can replace the DISK feature extractor with LiDAR systems, and Methods such as FPFH or SHOT will work well with our dual-head structure. The classification head identifies distinct areas in the point cloud data, whereas the similarity head compares point cloud segments. The strength of our method is its flexible design. The two-task approach works across different sensor types, because it focuses on the basic problem of finding when a robot returns to a known location.  Future studies will test these extensions more thoroughly and measure their performance against sensor-specific methods. We expect the benefits of few-shot learning and dual-task training to be carried over to these non-visual domains.
	
	\bibliographystyle{IEEEtran}
	\bibliography{LoopNet.bbl}
	
\end{document}